\documentclass[11pt]{article}

\usepackage[preprint]{acl}

\usepackage{times}
\usepackage{latexsym}
\usepackage{fvextra}

\usepackage[T1]{fontenc}

\usepackage[utf8]{inputenc}

\usepackage{microtype}

\usepackage{inconsolata}

\usepackage{graphicx}

\usepackage{booktabs}
\usepackage{tabularx}
\usepackage{amsmath, amssymb}
\usepackage{soul}
\usepackage{todonotes}
\usepackage{tikz}
\usetikzlibrary{positioning, shapes.geometric, calc}

\definecolor{queryBg}{HTML}{EBF3FE}  
\definecolor{replyBg}{HTML}{F4F4F5}  
\definecolor{queryText}{HTML}{1E3A8A} 
\definecolor{replyText}{HTML}{27272A} 

\title{A Language Model from 1913: Pretraining on Historical Text}

\author{
 \textbf{Xiaoxi Luo\textsuperscript{1,2}},
 \textbf{Zachary Shinnick\textsuperscript{3,4}},
 \textbf{Niclas Griesshaber\textsuperscript{4,5}},
 \textbf{Yixuan Wang\textsuperscript{1,2}},
\\
 \textbf{Junchi Yu\textsuperscript{4}},
 \textbf{Freda Shi\textsuperscript{1,2}},
 \textbf{Philip Torr\textsuperscript{4}},
 \textbf{Yao Lu\textsuperscript{6,}\thanks{Corresponding author.}}
\\
 \textsuperscript{1}University of Waterloo,
 \textsuperscript{2}Vector Institute,
 \textsuperscript{3}AIML, Adelaide University,
\\
 \textsuperscript{4}Department of Engineering Science, University of Oxford,
 \\
 \textsuperscript{5}Oxford Centre for Economic and Social History, University of Oxford
 \\
 \textsuperscript{6}Department of Computer Science, University College London
\\
\small\texttt{\{x25luo,tyxw,fhs\}@uwaterloo.ca}
 \quad
 \small\texttt{zachary.shinnick@adelaide.edu.au}
 \\
 \small\texttt{niclas.griesshaber@history.ox.ac.uk}
 \quad
 \small\texttt{\{junchi.yu,philip.torr\}@eng.ox.ac.uk}
 \quad 
 \small\texttt{yao.lu@cs.ucl.ac.uk}
}

\usepackage{color}

\begin{document}
\maketitle
\begin{abstract}
While modern language models increasingly rely on ever-larger web corpora, we show that pretraining on historical text (e.g., pre-1913 text) in a data-constrained setting can produce a temporally grounded language model that still shows reasonable performance on language understanding.
However, developing History LMs requires addressing challenges in data quality, preventing temporal leakage in post-training, and constructing temporally aligned evaluations.
We address these challenges and pretrain \textsc{TypewriterLM}, a 7.24B-parameter model with a 1913 knowledge cutoff.
We construct \textsc{TypewriterCorpus}, a 54B-token historical corpus with extensive temporal filtering, propose lexically grounded instruction tuning that constrains all responses to vocabulary from historical source documents, and introduce \textsc{History-Event}, a benchmark of 2,344 events for evaluating both competence and cutoff adherence.
We release \textsc{TypewriterLM} and all associated resources to support future research on History LMs.\footnote{Models and datasets are available at \href{https://huggingface.co/typewriter-ai}{hf.co/typewriter-ai}.}
\end{abstract}

\section{Introduction}

Modern foundation models \citep[\textit{inter alia}]{gemini-team-2025, qwen3, gpt2026, deepseekai-2026} achieve impressive performance across a wide range of natural language tasks. 
However, when used to study historical settings, they are biased toward modern stylistic and lexical distributions~\citep{underwood2025anachronism}.
Moreover, modern LMs have already been pretrained on the historical events and outcomes one might wish to analyze, so any apparent foresight could simply be memorized hindsight, which is known as lookahead bias~\citep{sarkar2024lookahead}. 

\begin{figure}[t]
\centering
\begin{tikzpicture}[
    node distance = 0.3cm,
    bubble/.style={
        rounded corners=6pt,
        inner xsep=10pt,       
        inner ysep=6pt,         
        align=flush left,
        font=\small\linespread{1.0}\selectfont
    },
    query/.style={
        bubble,
        text width=0.36\columnwidth, 
        fill=queryBg,
        text=queryText
    },
    reply/.style={
        bubble,
        text width=0.48\columnwidth, 
        fill=replyBg,
        text=replyText
    },
    header/.style={
        font=\footnotesize\bfseries,
        text=gray!80!black,
        inner sep=2pt
    }
]

    \node[header] (qHead) at (0,0) [anchor=west] {Query};
    \node[header] (rHead) at (\columnwidth,0) [anchor=east] {\textsc{TypewriterLM} Response};
    
    \node[query, below=of qHead.south west, anchor=north west] (q1) {
        Do you think there will soon be a major war in Europe?
    };
    \node[reply, yshift=-6pt, to path={}] (r1) at (q1.north -| rHead.east) [anchor=north east] {
        ``Certainly... It would not surprise me if this country were to go to war with France within ninety days.''
    };
    \node[query, below=12pt of q1.south west, anchor=north west] (q2) {
        Do you know about the general theory of relativity by Albert Einstein?
    };
    \node[reply, yshift=-6pt, to path={}] (r2) at (q2.north -| r1.north east) [anchor=north east] {
        ``...Professor A. Einstein, of whom we have not yet learned to speak as a physicist...''
    };
    \node[query, below=12pt of q2.south west, anchor=north west] (q3) {
        What is a computer?
    };
    \node[reply, yshift=-6pt, to path={}] (r3) at (q3.north -| r2.north east) [anchor=north east] {
        ``...an operator who multiplies or divides... a person who performs arithmetical operations...''
    };
\end{tikzpicture}
\caption{Probing \textsc{TypewriterLM} on ``future'' events.}
\label{fig:1913_probes}
\end{figure}

The cleanest fix is to train language models under a strict knowledge cutoff that excludes information beyond a chosen date, an approach that is attracting increasing attention from both machine learning and social sciences~\citep{london_llm_1800, goettlichetal2025, levine2026talkie}.
While modern LLMs benefit from massive and diverse web-collected corpora such as Common Crawl, only very limited historical data is available from OCR-transcribed sources.
This is a typical data-constrained pretraining setup~\citep{muennighoff-2023-scaling-data}.
For the post-training stage, instruction tuning typically relies on QA pairs from human annotators or frontier models.
For History LMs, neither route is viable, as each would inject the modern information and perspectives that the cutoff was meant to exclude. 
Finally, standard modern benchmarks are often temporally misaligned with History LMs, making it difficult to distinguish genuine reasoning limitations from unfamiliarity with modern language and contexts.

To address these challenges, we curate \textsc{TypewriterCorpus}, a 54-billion-token pretraining corpus collected from diverse sources and built with a strict data cleaning pipeline to mitigate temporal leakage. We introduce lexically grounded instruction tuning, where responses come directly from historical source documents, and construct \textsc{History-LIMA} and \textsc{History-SelfInstruct}. 
Using these resources, we train \textsc{TypewriterLM}, a 7.24B-parameter language model with a knowledge cutoff of 1913, in both base and instruction-tuned variants.

For evaluation, we consider both downstream capability and temporal consistency.
Our base model performs competitively on general benchmarks, and our instruction-tuned models are comparable with other History LMs trained with modern LLM supervision.
To evaluate temporal consistency, we construct \textsc{History-Event}, a benchmark of 2,344 significant historical events spanning 1700--2025. Using perplexity-based surprisingness evaluation, we find that History LMs become more surprised by post-cutoff events, whereas the modern baseline Llama-3.1-8B remains comparatively flat across time. 
This suggests that the intended historical cutoffs are meaningfully reflected in the models' learned knowledge distributions. 

To summarize, our contributions are as follows:

\begin{enumerate}
    \item We demonstrate that data-constrained pretraining can produce competitive, historically grounded language models, enabling future research in the humanities, social sciences and NLP.
    \item We design leakage-aware methods throughout the pipeline: strict cleaning for \textsc{TypewriterCorpus}; lexically grounded instruction tuning that sources every response from pre-cutoff documents (\textsc{History-LIMA} and \textsc{History-SelfInstruct}); and a historically aligned evaluation suite, \textsc{History-Event}, for detecting instances of data leakage.
    \item We openly release \textsc{TypewriterLM} (in both base and instruction-tuned versions) along with the full pipeline to support transparent research in this emerging area.
\end{enumerate}
\section{Related Work}

Research on using LLMs for temporal prediction tasks has focused on curating temporally bounded training datasets. 
This aims to prevent lookahead bias, where models are exposed to information from after the cutoff date during training. 
Since exposure to the test set biases a model's forecasting ability, efforts to mitigate lookahead bias have so far been primarily focused on financial applications. 
For example, families of language models have been trained on annual cutoff dates between 2000 and 2024 \citep{he2025chronologicallyconsistentlargelanguage,yan2026datedgptpreventinglookaheadbias,NBERw35247}.
These models are trained on approximately 100 billion tokens per year, with the largest containing 4 billion parameters. 
We address the same underlying challenge with a much earlier, ``historical'' cutoff date of 1913. This introduces additional challenges, yet we curate a pretraining corpus of more than 50 billion tokens to train a 7.24B-parameter model.

Even for modern LLMs, the reported and effective cutoff dates differ, emphasizing the difficulty of creating a training dataset with a well-defined cutoff date \citep{cheng2024dateddatatracingknowledge}. 
Even apparently bounded datasets, such as historical text corpora that can be scraped from online OCR-transcribed sources, may remain susceptible to similar leakage issues. 
For example, tailored OCR and layout detection pipelines have been used to build a large-scale historical newspaper dataset \citep{dell2023american}. This dataset was subsequently used to train a family of BERT-based 110M-parameter models with cutoff dates between 1900 and 1963 \citep{sarkar2024storieslm}.
Nonetheless, we construct our pretraining corpus using OCR text from institutional archives. Building text datasets directly from images remains difficult at scale due to the heterogeneity of historical documents, although recent benchmarking studies suggest that advances in vision-language models show promising progress toward a universal solution \citep{greif2025multimodalllmsocrocr,griesshaber2025multimodalllmshistoricaldataset}.

Recent efforts to scale History LMs to billions of parameters highlight the challenges of constructing training datasets without leakage. 
A 1.3B language model trained on London text from 1800--1875 still produces OCR artifacts such as ``Digitized by Google'' \citep{london_llm_1800}. This line of work was later scaled to 13B parameters using 260B pretraining tokens, but the resulting model exhibits temporal leakage, correctly identifying the U.S. president in 1936 despite a 1930 cutoff \citep{levine2026talkie}.
Other efforts to train History LMs include Ranke-4B, a family of language models introduced by a team of economic historians and trained on 80B tokens with cutoff dates between 1913 and 1946 \citep{goettlichetal2025}; a language model trained on Victorian-era British texts published between 1837 and 1899 \citep{MrChatterBoxd}; and a model trained on a pre-1900 text corpus \citep{Hla-gpt1900-2026}. With our data filtering approaches during pre- and post-training, we aim to further mitigate the risk of data leakage in this emerging area of research.
\section{Historical Pretraining} \label{pre-train}

\subsection{Corpus Construction}
\label{sec:data:sources}

We construct \textsc{TypewriterCorpus}, a historical English corpus spanning 1700--1913, 
where 1700 is the start of Late Modern English \citep{Barber_Beal_Shaw_2009}, and 1913 is the year immediately preceding World War I.
The corpus combines large-scale digitized books with curated historical datasets covering diverse domains, genres, and registers.

The majority of the corpus is derived from {Institutional Books} \citep{cargnelutti2025institutionalbooks10242b}, a large collection of digitized books from Harvard Library's collections spanning domains such as literature, science, law, and philosophy. 
We further incorporate {British Library Books} \citep[BL books;][]{BritishLibraryBooks2021}, another large-scale historical book collection digitized by the British Library.

To increase linguistic and stylistic diversity, we additionally include several curated historical corpora. 
Parliamentary discourse is represented by Hansard \citep{brezina2024hansard}; scientific writing by the Royal Society Corpus v6.0 \citep{fischer-etal-2020-royal}; legal and spoken language by Old Bailey \citep{huber-2016-old-bailey}; and literary Late Modern English by The Corpus of Late Modern English Texts v3.1 \citep[{CLMET;}][]{diller-2011-european-clmet}. 
We further include smaller corpora such as {Corpus of English Dialogues} \citep[{CED;}][]{kyto-2006-ced}, {The Lampeter Corpus of Early Modern English Tracts} \citep[{Lampeter;}][]{siemund-1997-lampeter}, {Corpus of Late Modern English Prose \citep[CLME Prose;][]{Denison-1994-CLMEP}, and a pamphlet collection \citep{bailyn_pamphlets_1750} to broaden coverage of dialogue, pamphlets, and prose styles.

To improve text quality, we apply several normalization and filtering procedures to remove OCR artifacts. 
For example, OCR systems frequently insert spaces at visual column or line boundaries (e.g., ``im- possible''), and we apply a conservative split-word joining pass. 
We also discard fragments with a high proportion of uppercase letters or symbols.
More details on \textsc{TypewriterCorpus} construction are provided in Appendix \ref{sec:appendix-pretrain-dataset}.

After data cleaning, \textsc{TypewriterCorpus} contains approximately 54 billion tokens as counted by our custom BPE tokenizer (\S \ref{sec:pretrain-tokenizer}). 
Table~\ref{tab:corpus} summarizes the corpus composition after filtering and cleaning. {Institutional Books} dominates the corpus by volume (97.7\%), while the remaining sources provide important diversity in genre, register, and domain. 
Figure~\ref{fig:token-year} illustrates the token distribution by decade, showing that the corpus is concentrated between 1800 and 1900.

\subsection{Leakage Mitigation}
\label{sec:data:leakage}

Temporal leakage presents a major challenge for History LMs, as archival documents frequently contain modern metadata or annotations. 
We identify several common sources of leakage and apply rule-based filtering to reduce contamination.

\paragraph{Institutional provenance metadata.}
Some scanned books contain ownership stamps (e.g., ``Harvard College Library''), institutional mottoes, catalog identifiers, and donor annotations added by modern libraries. We remove such provenance metadata from Institutional Books and related archival sources.

\paragraph{Web and HTML artifacts.}
URLs and HTML entities introduced during corpus processing are removed from all datasets.

\paragraph{Publishing and editorial metadata.}
Title-page imprint lines (e.g., book price, publisher addresses), attribution lines (e.g., shorthand-writer and editor credits in Old Bailey), and similar editorial metadata are removed heuristically.

Despite extensive filtering, completely eliminating temporal leakage from historical OCR corpora remains challenging. 
We also adopt leakage-aware post-training and evaluation protocols described in later sections.

\begin{table}[t]
  \centering
  \small
\setlength{\tabcolsep}{3pt}
  \begin{tabular}{l l r r}
    \toprule
    \textbf{Dataset} & \textbf{Domain / Genre} & \textbf{Tokens (M)} & \textbf{\%} \\
    \midrule
    Institutional & Books & 52,736 & 97.68 \\
    BL Books & Books & 915 & 1.69 \\
    Hansard & Parliamentary & 193 & 0.36 \\
    Royal Society & Scientific papers & 70 & 0.13 \\
    Old Bailey & Court proceedings & 39 & 0.07 \\
    CLMET & Literary & 31 & 0.06 \\
    CED & Dialogues & 0.67 & $<$0.01 \\
    Lampeter & Tracts & 0.35 & $<$0.01 \\
    CLME Prose & Prose & 0.12 & $<$0.01 \\
    Pamphlets & Pamphlets & 0.08 & $<$0.01 \\
    \midrule
    \textbf{Total} & & \textbf{53,986} & \textbf{100} \\
    \bottomrule
  \end{tabular}
  \caption{Composition of \textsc{TypewriterCorpus} after filtering and cleaning. Tokens are counted using our custom BPE tokenizer (\S \ref{sec:pretrain-tokenizer}).}
  \label{tab:corpus}
\end{table}

\begin{figure}
    \centering
    \includegraphics[width=\linewidth]{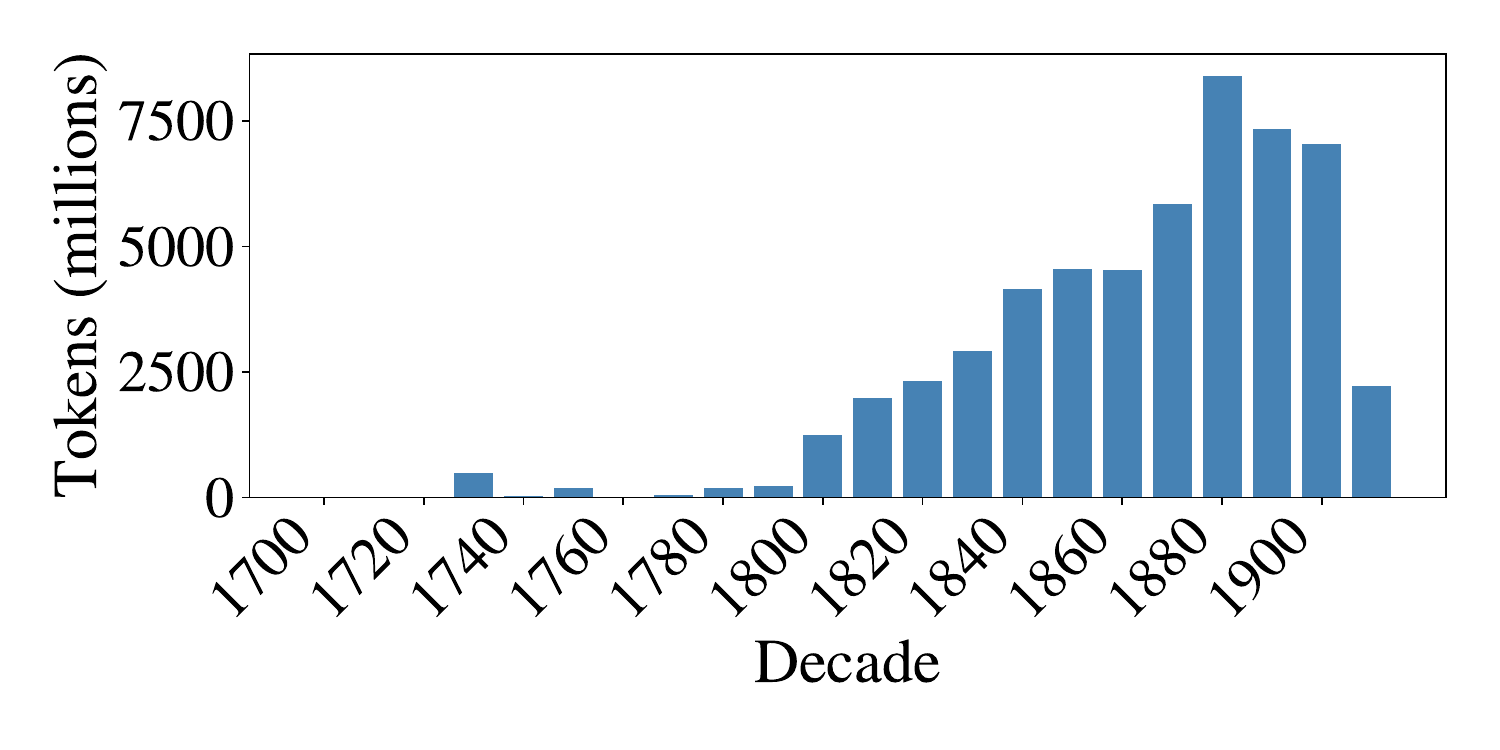}
    \caption{Number of tokens by decade (1700--1913) in our pretraining corpus. 
    }
    \label{fig:token-year}
\end{figure}

\subsection{Training Setup} \label{sec:pretrain-tokenizer}

To avoid modern vocabulary contamination, we train a custom byte-pair encoding (BPE) tokenizer~\citep{sennrich-etal-2016-bpe} with a vocabulary size of 32{,}000 on the training corpus, using the \texttt{o200k\_base} pre-tokenization strategy from \texttt{tiktoken}~\citep{OpenAI_tiktoken_2023}.

\textsc{TypewriterLM} is a 7.24B-parameter decoder-only Transformer following the Llama~3 architecture \citep{llama3-herd-models-2024}. The model uses grouped-query attention \citep{ainslie-etal-2023-gqa}, RMSNorm \citep{Zhang-2019-RMSnorm}, rotary position embeddings \citep{su2024roformer}, and an 8{,}192-token context length. 
We train using AdamW \citep{loshchilov-2018-adamw} with bfloat16 mixed precision on approximately 140B tokens (2.59 epochs over the corpus). 
More training setup details are provided in Appendix~\ref{appendix:training-detail}.
\section{Historical Instruction Tuning}

Instruction tuning can substantially shift the behaviour of pretrained language models~\citep{wei2022finetuned,ouyang2022training}. For History LMs, this presents a fundamental challenge: even when pretrained on temporally filtered corpora, standard supervised fine-tuning reintroduces modern linguistic and semantic distributions through modern instruction data or synthetic responses generated by modern LLMs. Filtering explicit factual references or simply prompting a modern LLM to imitate historical language is insufficient, since temporal leakage persists at the level of language distribution itself. 
While prior History LMs primarily focus on temporally filtering pretraining corpora~\citep{london_llm_1800,goettlichetal2025}, modern post-training pipelines, particularly reinforcement learning from AI feedback, inevitably introduce anachronistic linguistic and behavioural priors \citep{levine2026talkie}. 
In contrast, we explicitly address leakage introduced during instruction tuning itself. To this end, we construct the entire instruction-tuning corpus under a strict lexical grounding constraint: responses are constructed directly from pre-1913 source documents and may contain only lexical items appearing in the source passage, together with a small allowlist of function words.\footnote{E.g., closed-class grammatical tokens such as ``the'', ``this'', ``thou'', auxiliary verbs, conjunctions, and pronouns required for fluent composition.} 
Our goal is not merely to prevent factual anachronisms, but to preserve temporally consistent lexical, stylistic, and semantic patterns throughout instruction tuning.

\subsection{Dataset Curation}

\paragraph{Lexical Grounding Constraint.}
A response $y$ is accepted only if every lexical token is derivable from $V(p) \cup A$, where $V(p)$ denotes the set of words appearing in source text $p$ and $A$ is a small allowlist of function words; numerals must also appear in the source text. We enforce this constraint using a strict post-generation verifier that discards any non-compliant response. Verification is based primarily on exact lexical matching, with controlled recovery operations including limited morphological normalization, dehyphenation, page-break reconstruction, and fused-word correction to reduce false rejections caused by OCR artefacts. All recovery operations require the recovered form to remain derivable from the source text, preventing unsupported vocabulary while permitting limited lexical variation and OCR correction. 

\paragraph{\textsc{History-LIMA}.}
Following the ``less is more'' philosophy of \citet{zhou2023lima}, we construct a small high-quality instruction-tuning set consisting of 1,000 lexically grounded single-turn examples. Keeping the dataset small enables detailed human review while reducing drift away from the pretrained historical language distribution. Candidate pairs are generated using the lexical grounding approach described above and ranked using an LLM judge evaluating response quality and standalone usability, after which the top-scoring examples are retained. Instructions are then generated by Claude Opus to align with the grounded responses and manually reviewed by human annotators for historical coherence and knowledge leakage. We additionally construct a multi-turn variant containing 1,030 total examples. This extension includes 30 multi-turn dialogues following the conversational setup introduced in LIMA. Fifteen dialogues are curated from pre-1913 dialogue-oriented texts including catechisms and literary exchanges from authors such as Lucian and Dickens, while the remaining fifteen are derived from \emph{Hansard}~\citep{brezina2024hansard} parliamentary debates between 1853 and 1864, with assistant responses taken verbatim from the original speeches.

\paragraph{\textsc{History-SelfInstruct}.}
Inspired by Self-Instruct~\citep{wang2023self}, we scale instruction-tuning data by inverting the standard synthetic generation pipeline: responses remain fixed historical anchors, while only instructions are model-generated. The pipeline consists of three stages. \emph{(i)~Seed construction.} We use the 1,000 answer $\rightarrow$ question examples from \textsc{History-LIMA} as the seed set, preserving the same grounded-response design while treating instruction generation as the learned task. \emph{(ii)~Question-generator training.} We then fine-tune our pre-1913 base model on this seed set using LoRA ($r{=}64$) to obtain a generator that produces historically consistent instructions conditioned on arbitrary grounded responses. \emph{(iii)~Filtered generation at scale.} The generator produces candidate instructions for a large corpus of lexically grounded responses. Generated pairs are filtered using an LLM-judge to assess coherence, retaining 287{,}860 high-quality instruction-response pairs. Crucially, only the instructions are self-generated; all responses remain lexically grounded in historical source text.

\subsection{Training Details}
  \label{sec:itune-setup}
    We fine-tune the pre-1913 base model \textsc{TypewriterLM} separately on \textsc{History-LIMA} and \textsc{History-SelfInstruct} using parameter-efficient LoRA fine-tuning~\citep{hu2022lora}, yielding two instruction-tuned variants. Following~\citet{zhou2023lima}, the \textsc{TypewriterLM} (\textsc{LIMA}) model is trained for 15 epochs, while the \textsc{TypewriterLM} (\textsc{SelfInstruct}) model is trained for a single epoch. Our instruction-tuning setup is intentionally lightweight to minimize drift away from the pretrained historical language distribution while enabling instruction-following behaviours. Full training details and hyperparameters are provided in Appendix~\ref{app:it}.

\section{Evaluation}

Evaluating History LMs requires addressing two complementary questions. 
First, can they perform competitively on standard language understanding and reasoning benchmarks despite being trained under temporal and data constraints? 
Second, do they faithfully respect their intended historical cutoff?

To evaluate general capability, we test \textsc{TypewriterLM} on general benchmarks and compare it against several recent History LMs spanning different cutoff dates and model scales (\S \ref{sec:eval:general}): Mr.~Chatterbox \citep[0.34B, Victorian-era English]{MrChatterBoxd}, TimeCapsuleLLM-v2 \citep[1.22B, 1800--1875 English]{london_llm_1800}, GPT-1900 \citep[3.29B, pre-1900 English]{Hla-gpt1900-2026}, and Talkie-1930 \citep[13B, pre-1931 English]{levine2026talkie}. 
We additionally include GPT-2 XL \citep{radford-2019-language} and Qwen3-4B \citep{qwen3} for comparison.

To evaluate cutoff faithfulness, we further construct an evaluation suite measuring both temporal surprisingness and recall over historical events spanning 1700--2025 (\S \ref{sec:eval:surprisingness}). 
Together, these evaluations assess both downstream capability and preservation of historical knowledge boundaries.

\begin{table*}[t]
\centering
\small
\begin{tabular}{lccccc}
\toprule
\textbf{Model} & \textbf{Era} & \textbf{Size} & \textbf{ARC-E (\%)} & \textbf{ARC-C (\%)} & \textbf{HellaSwag (\%)} \\
\midrule
Mr.\ Chatterbox & 1837--1899 & 0.34B & 33.9 & 22.9 & 28.2  \\
TimeCapsuleLLM-v2 & 1800--1875 & 1.22B & 30.3 & 22.0 & 25.9 \\
GPT-1900 (base) & $\le$1900 & 3.29B & 36.7 & 23.9 & 32.3  \\
Talkie-1930 (base) & $\le$1930 & 13B & 55.8 & 35.6 & 49.8 \\
GPT-2 XL & $\le$2019 & 1.55B & 50.8 & 28.2 & 48.8 \\
Qwen3-4B (base) & $\le$2025 &4B & 77.4 & 51.6 & 65.9 \\
\midrule
\textsc{TypewriterLM} (base) & 1700--1913 & 7.24B & 43.3 & 27.9 & 36.1  \\
\bottomrule
\end{tabular}
\caption{
Accuracy (\%) on ARC-Easy (ARC-E), ARC-Challenge (ARC-C), and HellaSwag. All pairwise differences between \textsc{TypewriterLM} (base) and other models are statistically significant ($p<0.05$), except \textsc{TypewriterLM} (base) vs. GPT-2 XL on ARC-C. Details about significance tests are in Appendix \ref{app:significance}.
}
\label{tab:evaluation}
\end{table*}

\subsection{Knowledge Cutoff Adherence}
\label{sec:eval:surprisingness}

We evaluate whether \textsc{TypewriterLM} and other History LMs respect their knowledge cutoff by using two complementary metrics on \textsc{History-Event}, our evaluation dataset consisting of 2,344 historical events spanning the period 1700--2025. First, bits-per-byte (BPB) surprisingness provides a distribution-level signal of cutoff adherence. Second, recall provides an output-driven test for each event: pre-cutoff rates indicate factual correctness, while post-cutoff rates reveal data leakage.

\paragraph{Dataset.} We construct \textsc{History-Event} by scraping all events from the Wikipedia century timelines.\footnote{\url{https://en.wikipedia.org/wiki/Timeline_of_the_20th_century} and equivalent pages for the 18th, 19th, and 21st centuries. See~\ref{subsec:wikipedia-dataset}.} The scraped dataset contains the variables \texttt{event\_year} and \texttt{event\_description}. The number of events is distributed unevenly across centuries (Figure~\ref{fig:histevent_dist}).

\paragraph{Surprisingness.} Similar to \citet{levine2026talkie}, which draws on ideas from \citet{duderstadt_blogpost}, we measure how ``surprised'' a model is by providing it with the description of a historical event and the year in which it took place in the form: \begin{quote}\small \texttt{What do you think about the following event:} [\texttt{event\_description}]. \texttt{This took place in} [\texttt{event\_year}]. \end{quote}

The prefix serves as conditioning, while only the target span, i.e., the description and year phrase, is scored. Bits-per-Byte (BPB) surprisingness is calculated as 
\[
\text{BPB} = \frac{\text{NLL}_{\text{nats}}}{\ln 2 \cdot |\text{target}|_{\text{bytes}}}.
\]

where $\mathrm{NLL}_{\text{nats}}$ is the summed token negative log-likelihood of the target span under the model and $|\text{target}|_{\text{bytes}}$ is its length in bytes.

\paragraph{Recall.}

We further test whether a model can recall an event in \textsc{History-Event}. Depending on whether the event occurred before or after the cutoff date, a correct answer either implies that the model is factually correct or that it suffers from data leakage.\footnote{Even a correct pre-cutoff answer could stem from leaked post-cutoff sources (e.g., a modern account of a historical event). It may also be possible that a model has seen the historical event during its training, but cannot recall the event. See \S \ref{subsec:discussion-challenges}.} For each of the 2,344 historical events, we create a question in the following format:

\begin{quote}\small

\textit{``Do you know about the following event:} [\texttt{event\_description}]?
\textit{If so, explain what this event was and in what year did it take place?''}
\end{quote}

We exclude questions in which the \texttt{event\_description} contains a four-digit year. Gemini-3.1-Flash-Lite answers the remaining 2,148 questions. We keep 1,744 questions in which the response states the correct four-digit year from Wikipedia and demonstrates knowledge of the event. All History LMs answer the same 1,744 questions. Gemini-3.1-Flash-Lite then acts as an LLM-as-a-judge and receives four inputs: 1) the \texttt{event\_description} and 2) the \texttt{event\_year} from Wikipedia, 3) its previous gold answer, and 4) the History LM's answer. The judge scores the History LM's answer against the other three inputs under two criteria. Strict requires both the correct \texttt{event\_year} and correct information beyond simply rephrasing the \texttt{event\_description}, for which the gold answer serves as reference. Relaxed requires only the latter. Strict therefore provides a lower bound, while Relaxed provides an upper bound on recall. Since Gemini-3.1-Flash-Lite handles both selection and scoring, our results are biased toward historical events the modern model knows. Prompts are in Appendix~\ref{app:prompts}.

\begin{figure}
    \centering
    \includegraphics[width=\linewidth]{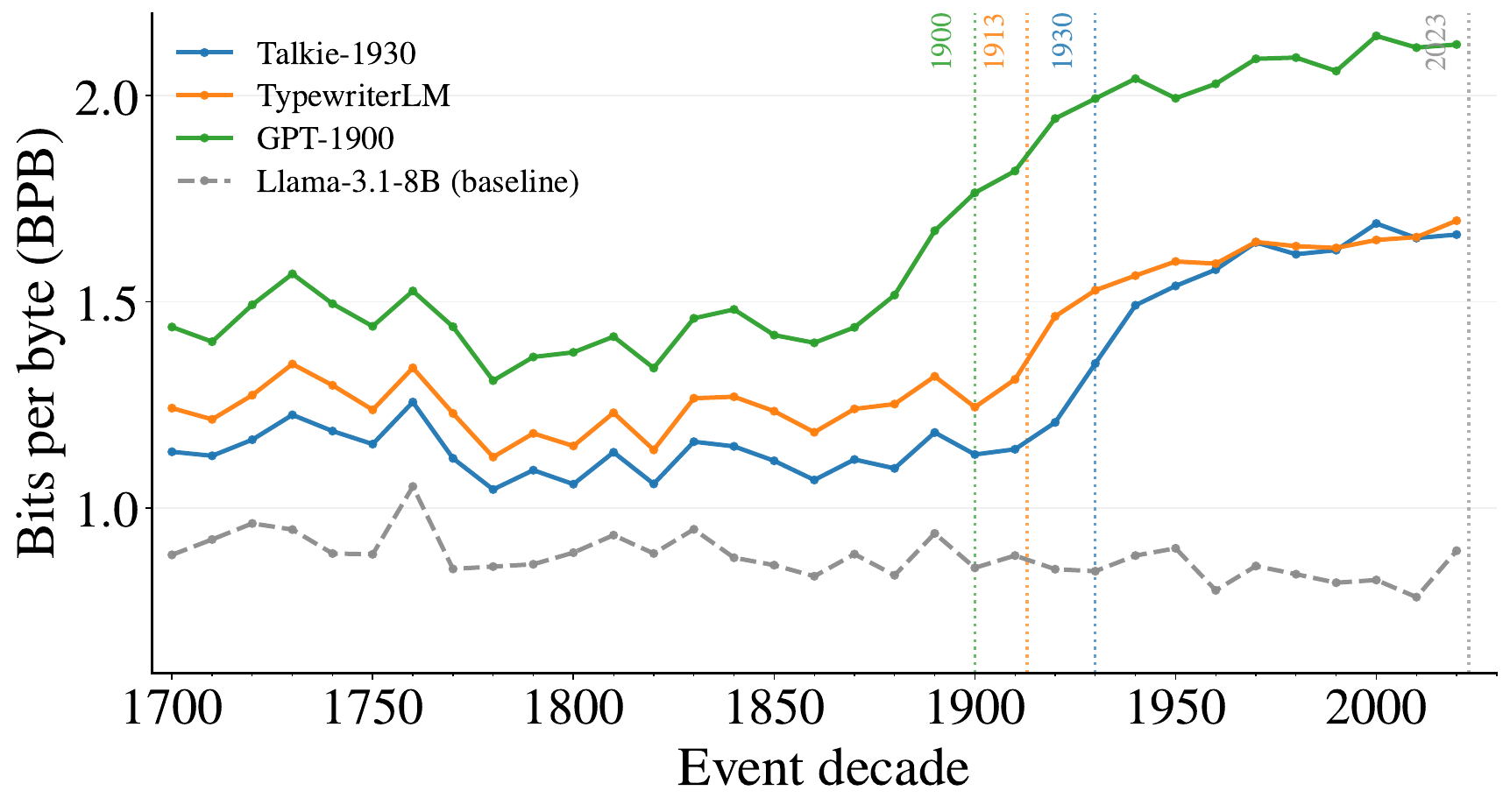}
    \caption{Bits-per-byte surprisingness on HISTORY-EVENT, averaged per decade, for three History LMs and a modern baseline. Dashed vertical lines mark the respective knowledge cutoff.}
    \label{fig:perplexity_headline_by_decade}
\end{figure}

\paragraph{Results.} Figure~\ref{fig:perplexity_headline_by_decade} depicts selected BPB surprisingness scores over time. All model statistics are in Appendix~\ref{app:bpb}. Table~\ref{tab:correctness_leakage} reports recall performance.

The BPB scores of the History LMs follow similar patterns, but begin to rise shortly before and after their respective cutoff dates, reaching a higher overall level in the post-cutoff period. The baseline, Llama-3.1-8B-Instruct \citep{llama3-herd-models-2024}, remains flat and even slightly declines for more recent periods, likely due to greater exposure to modern than historical data during training. Both patterns are expected, as the baseline was likely trained on all events, whereas the History LMs were not.

While the BPB scores provide evidence that the History LMs generally adhere to their respective knowledge cutoff dates, the metric is not sufficient to rule out data leakage. Therefore, we evaluate whether the models can recall each of the historical events. Crucially, using post-cutoff recall to measure leakage requires a History LM to first possess the capability to correctly recall pre-cutoff events.

Pre-cutoff, even the largest History LMs correctly recall only a modest fraction of events in \textsc{History-Event}, highlighting the difficulty of training capable History LMs.
TimeCapsuleLLM does not answer a single question correctly under the strict criterion. Mr. Chatterbox answers one (0.2\%). GPT-1900 (sft) answers 5.6\% strictly right (13.5\% relaxed). \textsc{TypewriterLM} (\textsc{Sel.In.}) reaches 6.2\% (20.3\% relaxed). Talkie-1930 (it) performs best, with 34.5\% strict and 53.3\% relaxed, with most gains coming after the instruction tuning.

Because only the largest History LMs demonstrate a solid pre-cutoff recall capability, we also find leakage primarily for Talkie-1930 and \textsc{TypewriterLM}, with up to 0.1\% of post-cutoff events leaking under the strict criterion and 0.9\% under the relaxed one (Table~\ref{tab:correctness_leakage}). 
Examples of data leakage are in Appendix~\ref{app:leakage}.

\begin{table*}[t]\centering\small
\begin{tabular}{llc cc cc cc}
\toprule
 & & & \multicolumn{2}{c}{Correctness (pre-cutoff)$\uparrow$} & \multicolumn{2}{c}{Leakage (post-cutoff)$\downarrow$} &
\multicolumn{2}{c}{$n$} \\
\cmidrule(lr){4-5}\cmidrule(lr){6-7}\cmidrule(lr){8-9}
Model & Size & Cutoff & Strict & Relaxed & Strict & Relaxed & pre & post \\
\midrule
Talkie-1930 (base) & 13B & 1930 & 2.9 & 8.8 & 0.0 & 0.3 & 719 & 1025 \\
Talkie-1930 (it) & 13B & 1930 & 34.5 & 53.3 & 0.0 & 0.9 & 719 & 1025 \\
GPT-1900 (sft) & 3.29B & 1900 & 5.6 & 13.5 & 0.0 & 0.2 & 498 & 1246 \\
Mr. Chatterbox & 340M & 1899 & 0.2 & 1.2 & 0.0 & 0.1 & 492 & 1252 \\
TimeCapsuleLLM & 1.2B & 1875 & 0.0 & 0.5 & 0.0 & 0.0 & 394 & 1350 \\
\midrule
\textsc{TypewriterLM} (base) & 7.24B & 1913 & 0.9 & 6.1 & 0.0 & 0.4 & 576 & 1168 \\  
\textsc{TypewriterLM} (\textsc{Sel.In.}) & 7.24B & 1913 & 6.2 & 20.3 & 0.1 & 0.6 & 576 & 1168 \\  
\textsc{TypewriterLM} (\textsc{LIMA}) & 7.24B & 1913 & 3.6 & 20.5 & 0.0 & 0.3 & 576 & 1168 \\
\bottomrule
\end{tabular}
\caption{Factual historical correctness (pre-cutoff, \%) and data leakage (post-cutoff, \%) on \textsc{History-Event}.}
\label{tab:correctness_leakage}
\end{table*}

\subsection{General Capability Benchmarks} \label{sec:eval:general}

\paragraph{Hellaswag.}
Standard HellaSwag \citep{zellers2019hellaswag} is a commonsense natural language inference benchmark written in modern English and derived from contemporary web text. 
We evaluate all aforementioned History LMs on the full validation set (10,042 examples), together with two modern baselines: GPT-2 XL and Qwen3-4B.
GPT-2 XL is an entry-level baseline in modern pretraining, yet remains non-trivial for History LMs because historical pretraining data cannot be scaled to the same extent as web corpora. Qwen3-4B, in contrast, represents a much stronger modern LLM and provides a reference point for performance achievable with large-scale modern pretraining.

As shown in Table~\ref{tab:evaluation}, performance on HellaSwag generally scales with model size among History LMs. 
However, Talkie-1930 requires substantially larger scale (13B parameters) to reach performance comparable to GPT-2 XL (1.55B), highlighting the difficulty of building competitive LMs from historical data alone. The gap may also partly reflect temporal and linguistic mismatch between History LMs and benchmarks constructed from contemporary web text.


\paragraph{ARC.}

The AI2 Reasoning Challenge \citep[ARC;][]{allenai-2018-arc} is a grade-school science multiple-choice benchmark. We evaluate on both ARC-Easy (2{,}376 questions) and ARC-Challenge (1{,}172 questions), using the template \texttt{``Question: \{\}\textbackslash nAnswer: ''} followed by each candidate answer.
As shown in Table~\ref{tab:evaluation}, \textsc{TypewriterLM} outperforms all smaller History LMs on both ARC subsets, but remains below Talkie-1930, GPT-2 XL and Qwen3-4B. 
Notably, \textsc{TypewriterLM} achieves performance on ARC-Challenge comparable to GPT-2 XL despite its historical cutoff, indicating its nontrivial reasoning capability.

\begin{table}[t]
\centering
\small
\begin{tabular}{lcc}
\toprule
\textbf{Model} & \textbf{Win Rate} & \textbf{Loss Rate} \\
\midrule
\textsc{TypewriterLM} (\textsc{LIMA})      & 46.5 & 53.5 \\
\textsc{TypewriterLM} (\textsc{Sel.In.})   & 42.0 & 58.0 \\
\bottomrule
\end{tabular}
\caption{
Pairwise AlpacaEval comparison against \textsc{GPT-1900} on 376 filtered prompts.
}
\label{tab:alpaca-eval}
\end{table}

\paragraph{AlpacaEval.}
Standard AlpacaEval prompts frequently require post-1913 knowledge and are therefore unsuitable for evaluating historically grounded language models. We therefore filter the 805 prompts from AlpacaEval 2.0~\citep{dubois2024length} using an LLM classifier, retaining only prompts that could plausibly be answered by a model trained exclusively on historical data. The resulting benchmark contains 376 historically valid prompts. We evaluate models using pairwise AlpacaEval comparisons judged by GPT-4o-mini following the AlpacaEval 2.0 protocol, including position-swapped comparisons to reduce positional bias.

Despite being fine-tuned under substantially stricter lexical grounding constraints, both \textsc{TypewriterLM} variants remain competitive against \textsc{GPT-1900}, with \textsc{TypewriterLM} (\textsc{LIMA}) achieving win rate 46.5\% (Table~\ref{tab:alpaca-eval}). 
Notably, \textsc{GPT-1900} is instruction-tuned using freely generated modern LLM responses, whereas our models constrain all assistant responses to lexically grounded historical source text. These results suggest that historically grounded instruction tuning can retain instruction-following behaviour without relying on unrestricted modern synthetic supervision.

\begin{table}[t]
\centering
\small
\setlength{\tabcolsep}{2.5pt}
\renewcommand{\arraystretch}{1.05}

\begin{tabular}{lcccc}
\toprule
\textbf{Model} &
\textbf{P-S} &
\textbf{P-L} &
\textbf{I-S} &
\textbf{I-L} \\
\midrule

GPT-1900
& 6.8
& 7.2
& 15.4
& 16.9 \\

Talkie-1930-IT
& 11.8
& 11.8
& 22.6
& 22.6 \\

\midrule

\textsc{TypewriterLM} (\textsc{LIMA})
& 4.6
& 5.3
& 13.5
& 14.2\\

\textsc{TypewriterLM} (\textsc{Sel.In.})
& 10.3
& 10.3
& 18.1
& 19.1 \\

\bottomrule
\end{tabular}

\caption{
IFEval Instruction-following accuracy (\%) under prompt-based (P) and instruction-based (I) settings using strict (S) and loose (L) matching. 
}
\label{tab:ifeval}
\end{table}

\paragraph{IFEval.}
We evaluate instruction-following using IFEval~\citep{zhou2023instruction}, a benchmark containing 541 prompts spanning 25 verifiable instruction types. 
Similar to AlpacaEval, we use an LLM-based classifier to filter for historically appropriate prompts, resulting in 263 prompts.
Following the standard protocol, we report prompt-level and instruction-level accuracy under both strict and loose matching criteria. 
Neither \textsc{TypewriterLM} variant differs significantly from GPT-1900 on any metric. 
Compared with Talkie-1930-IT, \textsc{TypewriterLM} (\textsc{SelfInstruct}) is not significantly different on three of the four metrics (except I-S), whereas \textsc{TypewriterLM} (\textsc{LIMA})---trained on only 1,030 examples---performs significantly worse on all four.
Despite substantially stricter lexical grounding constraints, \textsc{TypewriterLM} variants therefore achieve instruction-following performance broadly comparable to existing History LMs, which rely on frontier LLMs for synthetic instruction tuning or reinforcement learning from AI feedback during post-training. 
Details about significance tests are in Appendix \ref{app:significance}.

\section{Discussion: Challenges in History LM}
\label{subsec:discussion-challenges}

\paragraph{Challenges in Evaluation}
Modern benchmarks may fail to faithfully reflect the capabilities of History LMs due to two forms of temporal distribution shift: post-cutoff \emph{topics} and mismatched \emph{language style}. 
To disentangle these two effects, we construct two different versions of Hellaswag.
Starting from the original benchmark, we first apply keyword-based filtering to remove examples requiring post-1800 knowledge or cultural context, retaining 5{,}362 examples involving relatively timeless activities such as cooking, animal care, fishing, and family life. 
We then rewrite the filtered examples into a pre-1800 context using Claude Sonnet 4.6 while preserving the original task's domain and difficulty level. For example, ``baking cookies'' may be rewritten as ``baking biscuits at the hearth.'' 
This yields 2{,}048 multiple-choice examples written in 1800s prose style.

Table~\ref{tab:hellaswag1800} compares History LMs, GPT-2 XL, and Qwen3-4B on the original, filtered, and rewritten benchmarks. 
For all models except Talkie-1930, topic filtering yields insignificant changes. 
Rewriting has a substantially larger and significant effect for all models except Talkie-1930, increasing performance for the other History LMs while decreasing it for modern LMs.
These results suggest that benchmark performance is strongly influenced by temporal mismatch in language style.

\begin{table}[t]
\centering
\setlength{\tabcolsep}{3pt}
\small
\begin{tabular}{lccc}
\toprule
\textbf{Model} & \textbf{Original} & \textbf{Filtered} & \textbf{Rewritten} \\
\midrule
GPT-2 XL &   48.8  &   48.2   &   41.2\rlap{$^{***}$} \\
Qwen3-4B & 65.9 & 66.2 & 54.0\rlap{$^{***}$} \\
GPT-1900 (base) & 32.3 & 32.3 & 36.0\rlap{$^{**}$} \\
Talkie-1930 (base) & 49.8 & 51.2\rlap{$^{**}$} & 52.3 \\
\midrule
\textsc{TypewriterLM} (base) & 
36.1 & 36.2 & 40.0\rlap{$^{**}$} \\
\bottomrule
\end{tabular}
\caption{
Accuracy (\%) on the original HellaSwag validation set, the topic-filtered subset, and the rewritten subset. Stars on `Filtered' column: significance levels of `Original' vs.\ `Filtered'. Stars on `Rewritten' column: significance levels of `Filtered' vs.\ `Rewritten'. Details about significance tests are in Appendix \ref{app:significance}.
}
\label{tab:hellaswag1800}
\end{table}

\paragraph{Challenges in Identifying Leakage.}
Although we apply extensive filtering during both corpus construction and post-training, leakage remains difficult to eliminate completely, as demonstrated by our \textsc{History-Event} evaluation suite. We conduct several case studies and identify one source of leakage: subtle metadata artifacts. 
In our largest source, the Institutional dataset, the filtering pipeline uses each source's primary date. While this corresponds to the print year for monographs, for continuing resources such as journals and collections it may instead denote the year in which the series began. 
Post-cutoff volumes of long-running serials that started before 1913 therefore could pass our filtering. Appendix~\ref{app:leakage} traces two leaked responses, the 1918 foundation of Czecho-Slovakia and the 1964 Civil Rights Act, to source volumes affected by this artifact. Finally, the model may suffer from additional data leakage not captured by our evaluation suite, such as post-1913 social values, which is difficult to detect since doing so requires eliciting and judging specific responses.

Leakage-free History LMs are crucial for applications requiring strict temporal fidelity, such as the AGI Einstein test \citep{Perrigo2025} and historically grounded social science. Our work highlights the difficulty of developing a completely leakage-free data pipeline and we hope these findings draw more attention to this issue, particularly as this challenge is likely to be exacerbated at scale.

\section{Future Directions}
History LMs open up a range of research directions across NLP and related fields.

History LMs could advance studies in reasoning--memorization interplay \citep[\textit{inter alia}]{razeghi-etal-2022-impact, xie-etal-2025-memorization}.
Since they are trained exclusively on pre-cutoff corpora, their performance on modern reasoning benchmarks is less likely to arise from direct memorization of benchmark data. In contrast, recent work has shown that modern LLMs can exhibit substantial performance degradation under relatively small perturbations or reformulations of popular reasoning benchmarks \citep[\textit{inter alia}]{oren-2024-proving, zhang-etal-2024-examination-llm}, suggesting potential benchmark contamination or shortcut memorization effects.

They also provide a natural setting for studying temporal distribution shift \citep{wiles-2022-dist-shift}.
Unlike conventional domain adaptation settings, the shift is historically grounded and affects multiple levels simultaneously, including vocabulary, language style, social values, and world knowledge. 

Historical corpora are finite and non-renewable, making History LMs a useful testbed for studying scaling behavior under data-limited pretraining setting \citep{muennighoff-2023-scaling-data}.

From the perspective of the humanities and social sciences, they are a novel tool to study temporally grounded language, culture, and social values. 
In Appendix~\ref{app:semantic-drift}, we present a preliminary exploration in this direction.
We probe \textsc{TypewriterLM (LIMA)} with terms that underwent lexical semantic change during the 19th century due to industrialization \citep{pedrazzini-mcgillivray-2022-machines}, and find that the model can surface both earlier senses and meanings or usages associated with industrialization. Beyond historical linguistics, History LMs may enable a new, generative approach to Retrospective Forecasting, using in-context learning to see how the model reacts to ``future'' events.

\section*{Limitations}
We do not systematically study the effect of dataset composition or mixing ratios during pretraining, despite the highly imbalanced proportions occupied by different corpora. 

Our pretraining corpus is US- and UK-centric: Institutional Books, which accounts for 97.68\% of the corpus, is sourced from Harvard Library's collections and weighted toward US publications, while BL Books, Hansard, the Royal Society Corpus, and Old Bailey add UK coverage. Historical records from other English-speaking regions and non-English-language sources are not represented. Expanding geographic and multilingual coverage is an important direction for future work.

As discussed in \S \ref{subsec:discussion-challenges}, we identify one source of leakage: a specific metadata artifact. In future work, we will develop a more fine-grained filtering pipeline that accounts for such document-level metadata artifacts.

Our model reflects one possible 1913 worldview due to its training cutoff and may generate content considered offensive by modern standards. As History LMs scale and become more capable, they may pose societal risks. Our future public releases will therefore include safety warnings and guardrails to mitigate harmful outputs.

\section*{Acknowledgements}
We are grateful for inspiring conversations with Daniel Göttlich. 

This research is supported in part by a Canada CIFAR AI Chair Award to Freda Shi.
Philip Torr is supported by an AI2050 Senior Fellowship from Schmidt Sciences.
This research is supported by the Engineering and Physical Sciences Research Council [grant number UKRI5028] to Yao Lu.
Research reported in this publication is supported by an Amazon Research Award, 2025, to Yao Lu. 
Niclas Griesshaber is highly grateful for the DPhil funding provided by the Economic and Social Research Council through UK Research and Innovation under the Advanced Quantitative Methods Award [grant number
ES/Y001761/1].

The authors acknowledge the use of resources provided by the Isambard-AI National AI Research Resource (AIRR). Isambard-AI is operated by the University of Bristol and is funded by the UK Government's Department for Science, Innovation and Technology (DSIT) via UK Research and Innovation; and the Science and Technology Facilities Council [ST/AIRR/I-A-I/1023].

\bibliography{custom}
\newpage
\appendix
\section{Pretraining}
\subsection{Pretraining Datasets} \label{sec:appendix-pretrain-dataset}
Here we provide a more detailed introduction to the datasets comprising \textsc{typewritercorpus}, and dataset-specific curation details.

\paragraph{Institutional Books.}
It is our largest source consisting of digitized books from Harvard Library's collections \citep{cargnelutti2025institutionalbooks10242b}, spanning 20 topics, e.g., literature, science, law, philosophy.
To guarantee data quality, we apply an OCR score filter, retaining only books with both original and post processed OCR scores higher than 92.
To avoid OCR fragments, We also discard any remaining paragraph shorter than 100 characters.
After filtering, this source contributes 52.74B tokens, 97.7\% of the total corpus.

\paragraph{British Library Books (BL Books).}
This dataset consists of books digitized by the British Library, covering a wide range of subject areas, and the majority were published in the 18--19th Century~\citep{BritishLibraryBooks2021}. 
This dataset provides page-level OCR score, and we only keep pages with OCR scores exceeding 80.
It contributes 914.66 M tokens after filtering.

\paragraph{Hansard.}
This dataset contains the official records of parliamentary proceedings and debates across the United Kingdom's legislative bodies, known as Hansard~\citep{brezina2024hansard}.
It represents a comprehensive collection of parliamentary discourse since 1802.
After data cleaning, it contributes 193.05 M tokens.

\paragraph{Royal Society Corpus v6.0.}
It is a diachronic corpus of scientific English spanning more than 300 years of scientific writing (1665–1996). It contains primarily scientific articles, derived from publications of the Royal Society of London \citep{fischer-etal-2020-royal}.
It contributes 70.38 M tokens.

\paragraph{Old Bailey Corpus.}
This corpus captures speech-related uses of Late Modern English in London's Central Criminal Court \citep{huber-2016-old-bailey}.
It contributes 38.51 M tokens.

\paragraph{The Corpus of Late Modern English Texts v3.1 (CLMET).}
The corpus covers the period 1710--1920, covering five major genres: narrative fiction, narrative non-fiction, drama, letters and treatise, and unclassified texts \citep{diller-2011-european-clmet}. 
After filtering to 1913, it contributes 31.46 M tokens.

\paragraph{Corpus of English Dialogues (CED).}
This corpus contains dialogues from literary and didactic works from 1560 to 1760 \citep{kyto-2006-ced}.
After filtering, it contributes {0.67} million tokens.

\paragraph{The Lampeter Corpus of Early Modern English Tracts (Lampeter).}
The Lampeter Corpus comprises political, economic, and religious pamphlets and tracts from 1640--1740~\citep{siemund-1997-lampeter}.
After filtering, it contribute 0.35 million tokens.

\paragraph{Corpus of Late Modern English Prose (CLME Prose).}
This corpus provides selected prose texts from the late Middle and early Modern English periods \citep{Denison-1994-CLMEP}.
After filtering to the target window, it contributes {0.12} million tokens.

\paragraph{Pamphlets.}
This corpus contains curated set of five historical pamphlets \citep{bailyn_pamphlets_1750} in clean digitized form, and contributes 0.08 million tokens.

\subsection{Training Details} \label{appendix:training-detail}
Our tokenizer adds \texttt{<bos>} before a document and appends \texttt{<eos>} after it.

Our base model is a 7.24B-parameter decoder-only Transformer following the Llama 3 architecture~\citep{llama3-herd-models-2024}. 
It has 32 layers, hidden size 4096, and SwiGLU feed-forward blocks with inner dimension 14336~\citep{shazeer2020gluvariantsimprovetransformer}. 
Attention uses 32 query heads with grouped-query attention \citep[GQA;][]{ainslie-etal-2023-gqa} sharing 8 key/value heads, with head dimension 128. 
We use RMSNorm~\citep{Zhang-2019-RMSnorm}, rotary position embeddings~\citep{su2024roformer}, and a context length of 8192 tokens. 

We optimize with AdamW~\citep{loshchilov-2018-adamw} ($\beta_1{=}0.9$, $\beta_2{=}0.95$, weight decay 0.1) and gradient clipping at $1.0$. 
The learning rate follows a cosine schedule with peak $3{\times}10^{-4}$, minimum $3{\times}10^{-5}$.
We use bfloat16 mixed precision for training.
Pretraining uses approximately 140~B tokens, approximately $2.59$ epochs over the training corpus.

\section{Instruction Tuning}
More details about instruction tuning can be found in Table \ref{tab:itune-hparams}. 
\label{app:it}
  \begin{table}[h]
  \centering
  \small
  \scalebox{0.9}{
  \begin{tabular}{lcc}
  \toprule
                         & \textsc{Hist.-LIMA} & \textsc{Hist.-SelfInstr.} \\
  \midrule
  Training pairs         & 1{,}030             & 287{,}860                  \\
  Epochs                 & 15                  & 1                          \\
  LoRA $r$ / $\alpha$    & 64 / 128            & 64 / 128                   \\
  LoRA dropout           & 0.05                & 0.05                       \\
  Trainable params       & 430\,M (5.6\%)      & 430\,M (5.6\%)             \\
  Learning rate          & $1{\times}10^{-4}$  & $1{\times}10^{-4}$         \\
  Schedule               & cosine, 3\% warmup  & cosine, 3\% warmup         \\
  Weight decay           & 0                   & 0                          \\
  Effective batch        & 16                  & 64                         \\
  Hardware               & 1$\times$GH200      & 4$\times$GH200             \\
  Max seq.\ length       & 2048                & 2048                       \\
  Optimiser              & AdamW               & AdamW                      \\
  Precision              & bf16                & bf16                       \\
  \bottomrule
  \end{tabular}
  }
  \caption{LoRA SFT hyperparameters. Both adapters target all attention and MLP projection matrices and additionally fine-tune the token embedding and LM head to accommodate ChatML special tokens.}
  \label{tab:itune-hparams}
  \end{table}

\section{Evaluation Details}
\label{app:eval}

\subsection{Prompts}
\label{app:prompts}

Each History LM answer is judged under the strict and relaxed criterion:

\paragraph{Judge: Strict Criterion.}
\begin{Verbatim}[breaklines=true, frame=single, fontsize=\footnotesize, commandchars=\\\{\}]
You score a model's answer to a historical-event question against ground truth. The model was SHOWN the event description in the question, so simply restating or paraphrasing that description is NOT evidence of knowledge. You are given the event description, the CORRECT YEAR (from Wikipedia), a REFERENCE answer, and the MODEL's answer.

Award score = 1 if and only if BOTH hold: (a) the answer states the correct year ({event_year}); and (b) it provides correct, specific information about the event that GOES BEYOND the description (e.g., the outcome, consequences, named participants, or the date), consistent with the reference -- not a restatement of the description and not a plausible-sounding guess. Award score = 0 otherwise.

Return ONLY valid JSON with no markdown fences:
{"score": 0 or 1, "reasoning": "<one sentence>"}

---
Event description (shown to the model): {event_description}
Correct year (ground truth): {event_year}
Reference answer: {gold_answer}

Model's answer:
{model_answer}
\end{Verbatim}

\paragraph{Judge: Relaxed Criterion.}

\begin{Verbatim}[breaklines=true, frame=single, fontsize=\footnotesize, commandchars=\\\{\}]
You judge whether a model's answer shows knowledge of a historical event BEYOND what it was told. The model was SHOWN the event description in the question, so merely restating or paraphrasing that description is NOT evidence of knowledge. You are given the event description, the correct year (from Wikipedia), a REFERENCE answer, and the MODEL's answer.

Award score = 1 if the answer contains at least some correct, specific information about the event that GOES BEYOND the description (e.g., the date, outcome, consequences, or named participants not already stated in the description), consistent with the reference. Award score = 0 if the answer only restates or paraphrases the description, is vague, is wrong, or hallucinates.

Return ONLY valid JSON with no markdown fences:
{"score": 0 or 1, "reasoning": "<one sentence>"}

---
Event description (shown to the model): {event_description}
Correct year (ground truth): {event_year}
Reference answer: {gold_answer}

Model's answer:
{model_answer}
\end{Verbatim}

\subsection{Dataset and Filtering}
\label{subsec:wikipedia-dataset}

We scraped the 2,344 historical events from the following four Wikipedia pages, accessed in May 2026:

\begin{itemize}
    \item \url{https://en.wikipedia.org/wiki/Timeline_of_the_18th_century}
    \item \url{https://en.wikipedia.org/wiki/Timeline_of_the_19th_century}
    \item \url{https://en.wikipedia.org/wiki/Timeline_of_the_20th_century}
    \item \url{https://en.wikipedia.org/wiki/Timeline_of_the_21st_century}
\end{itemize}

We filter them as follows:

\label{app:dataset}
\begin{enumerate}\itemsep1pt
  \item Remove events whose description contains a year, so it cannot be copied from the question:
        $2{,}344 \rightarrow 2{,}148$ ($-196$).
  \item Keep only events for which Gemini-3.1-Flash-Lite's answer states the correct year from Wikipedia and demonstrates knowledge of the event: $\rightarrow \mathbf{1{,}744}$ historical events are retained.
\end{enumerate}

\begin{figure}[htbp!]\centering
  \includegraphics[width=\linewidth]{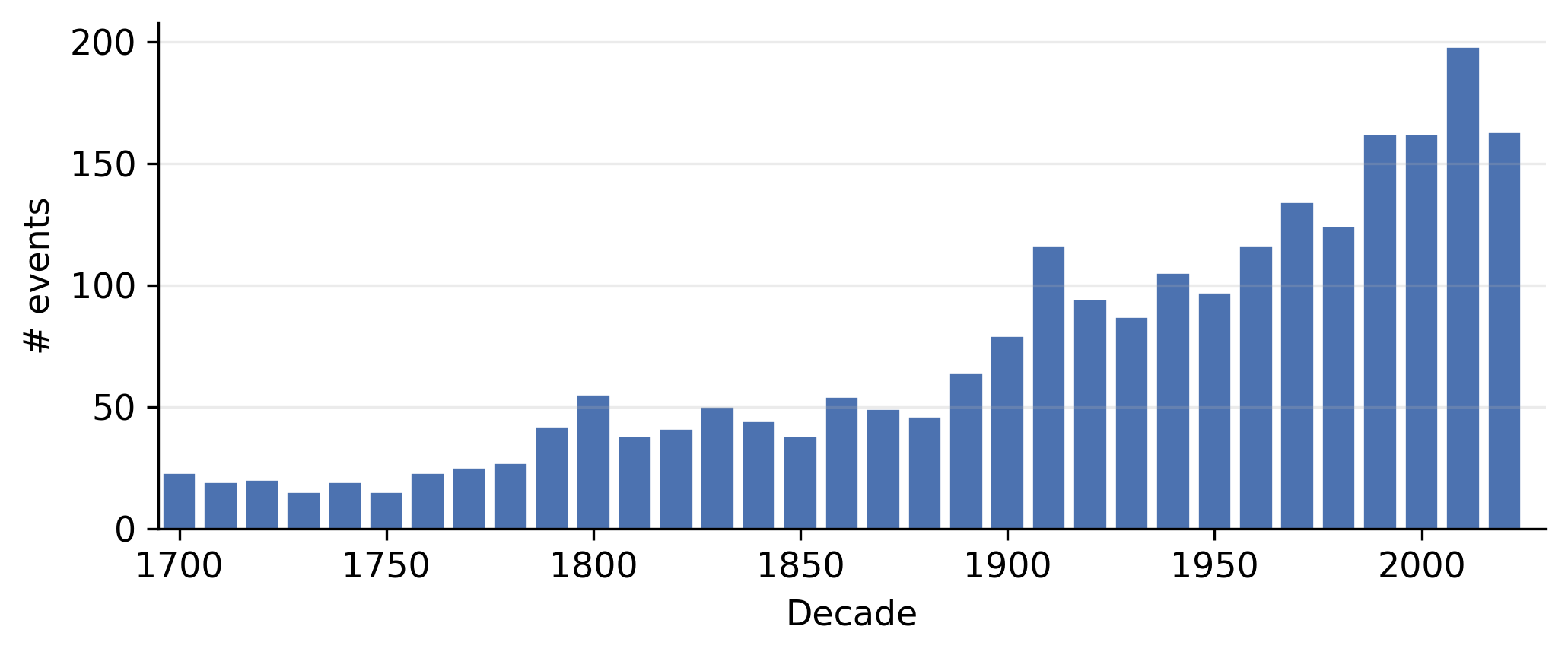}
  \caption{Number of historical events in \textsc{History-Event} per decade (1700--2025).}
  \label{fig:histevent_dist}
\end{figure}

\subsection{Full BPB Statistics}
\label{app:bpb}
The full BPB statistics are provided in Table \ref{tab:bpb_full}. Llama-3.1-8B-Instruct (our modern baseline) is included for reference, but only includes 38 events post-cutoff.

\begin{table*}[t]\centering\small
\begin{tabular}{llcccc}
\toprule
Model & Size & Cut. & BPB$_{\text{pre}}$ & BPB$_{\text{post}}$ & Ratio \\
\midrule
Llama-3.1-8B-Instruct                      & 8B    & 2023 & 0.855 & 1.214 & 1.42 \\
\midrule
Talkie-1930 (base)                         & 13B   & 1930 & 1.138 & 1.605 & 1.41 \\
Talkie-1930 (it)                           & 13B   & 1930 & 1.320 & 1.870 & 1.42 \\
\textsc{TypewriterLM} (base-v2)            & 7.24B & 1913 & 1.235 & 1.604 & 1.30 \\  
\textsc{TypewriterLM} (v2-selfinstruct)    & 7.24B & 1913 & 1.227 & 1.616 & 1.32 \\  
\textsc{TypewriterLM} (v2-LIMA)            & 7.24B & 1913 & 1.958 & 2.540 & 1.30 \\
GPT-1900 (base)                            & 3.29B & 1900 & 1.455 & 2.038 & 1.40 \\
GPT-1900 (sft)                             & 3.29B & 1900 & 1.541 & 2.186 & 1.42 \\
Mr.\ Chatterbox                            & 340M  & 1899 & 1.671 & 2.254 & 1.35 \\
TimeCapsuleLLM                             & 1.2B  & 1875 & 1.650 & 1.960 & 1.19 \\
\bottomrule
\end{tabular}
\caption{Bits-per-byte before/after each model's cutoff. Ratio = $\frac{\text{BPB}_{\text{post}}}{\text{BPB}_{\text{pre}}}$}
\label{tab:bpb_full}
\end{table*}

\subsection{Models}
\label{app:models}
An overview of the models we used is in Table \ref{tab:models}.

\begin{table*}[t]\centering\small
\begin{tabular}{lll}
\toprule
Model & Hugging Face ID & Cut. \\
\midrule
talkie-1930-13b-base   & \texttt{talkie-lm/talkie-1930-13b-base}      & 1930 \\
talkie-1930-13b-it    & \texttt{talkie-lm/talkie-1930-13b-it}        & 1930 \\
\textsc{TypewriterLM} (base-v2)           & \texttt{typewriter-ai/typewriter-1913-7B-base-v2} & 1913 \\
\textsc{TypewriterLM} (v2-selfinstruct) & \texttt{typewriter-ai/typewriter-1913-v2-selfinstruct}  & 1913 \\
\textsc{TypewriterLM} (v2-sft)          & \texttt{typewriter-ai/typewriter-1913-7B-sft-v2} & 1913 \\
GPT-1900 (base)     & \texttt{mhla/gpt1900-d34-22btok}             & 1900 \\
GPT-1900 (sft)      & \texttt{mhla/gpt1900-instruct-v3-sft}        & 1900 \\
Mr.\ Chatterbox     & \texttt{tventurella/mr\_chatterbox\_model}   & 1899 \\
TimeCapsuleLLM      & \texttt{haykgrigorian/TimeCapsuleLLM-v2-llama-1.2B} & 1875 \\
Llama-3.1-8B-Instruct & \texttt{meta-llama/Llama-3.1-8B-Instruct}    & 2023 \\
Gemini-3.1-flash-lite       & -               & 2025 \\
GPT-2 XL & \texttt{openai-community/gpt2-xl} & 2019 \\
Qwen3-4B & \texttt{Qwen/Qwen3-4B} & 2025\\
\bottomrule
\end{tabular}
\caption{Overview of the evaluated models}
\label{tab:models}
\end{table*}

\subsection{Post-Cutoff Leakage Cases}
\label{app:leakage}

Below, we present two instances of data leakage traced back to our training dataset, which were identified via our \texttt{History-Event} evaluation suite and manual exploration.

\noindent \textbf{\textsc{TypewriterLM}, The Republic of Czecho-Slovakia, 1918 (cutoff 1913).}\\
\textbf{Q:} The Republic of Czecho-Slovakia was founded in which year?\\
\textbf{A:} Czechoslovakia was proclaimed a republic on November 27, 1918. \\
\textbf{Potential source}:\\
\textit{``The Republic of Czecho-Slovakia, declared October 18, 1918, includes the former Austrian provinces of Bohemia, Moravia, and a part of Silesia, as well as territories formerly under the Hungarian [\dots]''}
\begin{flushright}
\textit{---Mineral resources of the United States}\\(started from 1882)
\end{flushright}

\noindent \textbf{\textsc{TypewriterLM}, Civil Rights Act (v1 model; not exactly reproduced by v2), 1964 (cutoff 1913).}\\
\textbf{Q:} \emph{Do you know about the following event: `Civil Rights Act abolishes segregation in the USA'? If so, explain what this event was and in what year did it take place?} \\
\textbf{A:} ``The Civil Rights Act of 1964 abolished segregation in the USA and expanded freedom
of choice by eliminating barriers to access to many of the social services and educational
facilities previously prohibited. [\dots]''\\
\textbf{Potential source}:\\
\textit{``King, Jr. and those that joined him in the civil rights movement resulted in the enactment of the Civil Rights Act of 1964, the Voting Rights Act of 1965, and the Fair Housing [\dots]''}
\begin{flushright}
\textit{---Journal of the House of Representatives\\of the United States (started from 1789)}
\end{flushright}

\noindent \textbf{\textsc{TypewriterLM} (Self-Instruct), International Labour Organization, 1919 (cutoff 1913).}\\
\textbf{Q:} \emph{Do you know about the following event: `The International Labour Organization is established'? If so, explain what this event was and in what year did it take place?} \\
\textbf{A:} ``The International Labour Organization was established at a Conference of Nations held at Washington in 1919.''\\
\textbf{Potential source}: not yet traced in \textsc{TypewriterCorpus}.

\noindent \textbf{Talkie 1930 (it), New Deal (previous run), 1933 (cutoff 1930).}\\
\textbf{Q:} \emph{Do you know about the following event: `New Deal begins in America'? If so,
explain what this event was and in what year did it take place?}\\
\textbf{A:} ``In 1933, the New Deal began in the United States when Congress passed a bill establishing a national economic council to plan social and economic reconstruction. [\dots]''\\

\section{Lexical Semantic Change Cases}
\label{app:semantic-drift}
Following \citet{pedrazzini-mcgillivray-2022-machines}, we qualitatively probe words that underwent semantic changes during the Industrial Revolution, which falls within \textsc{TypewriterLM}'s 1700--1913 training period.
We prompt \textsc{TypewriterLM} (\textsc{LIMA}) with ``What is the meaning of the word `\texttt{[word]}'?'' to examine which senses are most readily elicited.

Specifically, we examine nine terms discussed by \citet{pedrazzini-mcgillivray-2022-machines}: \emph{train}, \emph{wheel}, \emph{traffic}, \emph{gear}, \emph{coach}, \emph{stamp}, \emph{matches}, \emph{fellow}, and \emph{railway}. 
We observe a mixed pattern---for some words, \textsc{TypewriterLM} produces meanings associated with industrialization, and for other words, the same prompt elicits earlier senses.

\subsection{Senses Related to Industrialization}

\begin{quote}
\textbf{train:} A train is a body of cars, either passenger or freight, drawn by a locomotive. A passenger train is generally divided into two or more passenger coaches, which are coupled together, and draw the train at the rate of from twenty to forty miles an hour...
\end{quote}

\begin{quote}
\textbf{wheel:} A wheel is a circular disk, revolved by a locomotive, and serving the purpose of a carriage to transport passengers or freight over a track specially provided for that purpose.
\end{quote}

\begin{quote}
\textbf{traffic:} Traffic is the act of moving goods or persons, or ideas; commerce; traffic. The word is now chiefly used in regard to matters of trade, of which transportation is a part...
\end{quote}

\begin{quote}
\textbf{gear:} Gear is a noun; and, as such, applies only to the things used in connection with the locomotive and cars, to produce motion. Among these are the wheels and axles...
\end{quote}

\subsection{Earlier Senses}
\begin{quote}
\textbf{coach:} A coach is a covered vehicle for the conveyance of passengers, standing on two wheels, and drawn by two horses.
\end{quote}

Here, the model favours the horse-drawn vehicle sense rather than the later railway-carriage sense.

\begin{quote}
\textbf{stamp:} A word in general use, denoting the act of stamping, or the act of forcing so as to fix the seal.
\end{quote}

The response reflects the older sense related to marking or imprinting.

\begin{quote}
\textbf{matches:} A word now in general use, denoting the act of matching or adjusting any thing in competition...
\end{quote}

The model does not produce the later sense referring to phosphorus matches.

\begin{quote}
\textbf{fellow:} The word fellow is a noun, and, as such, applies only to two or more persons considered in relation to each other...
\end{quote}

The response emphasizes an older relational sense rather than some of the broader modern usages.

\begin{quote}
\textbf{railway:} An iron railway or track, on which are fixed rails for the wheels of carriages to run on.
\end{quote}

The model favors the physical-track sense rather than broader modern meanings referring to passenger transportation systems or railway organizations.

Overall, \textsc{TypewriterLM} can surface lexical senses from different periods within its training range, rather than uniformly favouring either archaic or contemporary meanings. 
Yet, we acknowledge that these examples should be interpreted as a preliminary exploration, instead of a complete characterization of the model's lexical knowledge: a polysemous word may be associated with multiple senses, and which one is produced can depend on the prompt and context. 

\section{Statistical Significance Testing}
\label{app:significance}

\paragraph{Table~\ref{tab:evaluation}.} We compare each baseline against
\textsc{TypewriterLM} on the same examples with a two-sided McNemar's test.

\paragraph{Table~\ref{tab:ifeval}.} We compare each \textsc{TypewriterLM} variant
against Talkie-1930-IT and GPT-1900 on the same 263 filtered prompts, using a
two-sided paired McNemar's test. 

\paragraph{Table~\ref{tab:hellaswag1800}.} The filtered set is a subset of the original benchmark, and the rewritten set a further subset of the filtered examples rewritten into period-appropriate prose.
For the \textbf{original vs.\ filtered} effect, we compare the 4{,}680 excluded examples against the 5{,}362 retained examples using a two-sided $z$-test, as these are disjoint sets of examples.
For the \textbf{filtered vs.\ rewritten} effect, we conduct a one-sided paired McNemar's test on the 2,048 rewritten items, comparing their filtered and rewritten form.

\end{document}